\documentclass[10pt,twocolumn,letterpaper]{article}

\usepackage{btas}
\usepackage{times}
\usepackage{epsfig}
\usepackage{graphicx}
\usepackage{amsmath}
\usepackage{amssymb}

\usepackage{multirow}
\usepackage{textcomp}
\usepackage{array}

\graphicspath{{./graphics/}}
\usepackage{url}

\btasfinalcopy 

\usepackage{fancyhdr}
\begin{document}

\title{Assessment of iris recognition reliability for eyes affected by ocular pathologies}

\author{Mateusz Trokielewicz$^{\dag,\ddag}$, Adam Czajka$^{\ddag,\dag}$\\
$^{\dag}$Biometrics Laboratory\\
Research and Academic Computer Network\\
Wawozowa 18, 02-796 Warsaw, Poland\\
$^{\ddag}$Institute of Control and Computation Engineering\\
Warsaw University of Technology\\
Nowowiejska 15/19, 00-665 Warsaw, Poland\\
{\tt\small {mateusz.trokielewicz,adam.czajka}@nask.pl}
\and
Piotr Maciejewicz$^{\star}$\\
$^{\star}$Department of Ophthalmology\\
Medical University of Warsaw\\
Lindleya 4, 02-005 Warsaw, Poland\\
{\tt\small piotr.maciejewicz@wum.edu.pl}
}

\maketitle
\thispagestyle{empty}

\begin{abstract}

This paper presents an analysis of how the iris recognition is impacted by eye diseases and an appropriate dataset comprising 2996 iris images of 230 distinct eyes (including 184 illness-affected eyes representing more than 20 different eye conditions). The images were collected in near infrared and visible light during a routine ophthalmological practice. The experimental study shows four valuable results. First, the enrollment process is highly sensitive to those eye conditions that make the iris obstructed or introduce geometrical distortions. Second, even those conditions that do not produce visible changes to the iris structure may increase the dissimilarity among samples of the same eyes. Third, eye conditions affecting iris geometry, its tissue structure or producing obstructions significantly decrease the iris recognition reliability. Fourth, for eyes afflicted by a disease, the most prominent effect of the disease on iris recognition is to cause segmentation errors. To our knowledge this is the first database of iris images for disease-affected eyes made publicly available to researchers, and the most comprehensive study of what we can expect when the iris recognition is deployed for non-healthy eyes.

\end{abstract}

\section{Introduction}

Intricate and unique iris patterns provide attractive biometric features that, when calculated for healthy eyes, allow to build large biometric applications, such as India's AADHAAR\footnote{\url{https://uidai.gov.in}} with more than 600 million people enrolled up to now, or the CANPASS\footnote{\url{http://www.cbsa- asfc.gc.ca/prog/canpass/canpassair-eng.html}} system maintained by Canadian Border Services Agency (CBSA) and providing an efficient entry into Canada for frequent travelers. However, there are various medical conditions affecting the eye structures, in particular the iris, and possibly deteriorating the recognition reliability. When investigating past work (Sec. \ref{sec:RelatedWork}) it is clear that due to the lack of large, heterogeneous, and publicly available databases appropriate to this subject, we are still far from full understanding of how various eye conditions impact iris recognition. This paper aims at answering four crucial questions related to eye conditions and their impact on iris recognition, and delivers an appropriate database of non-healthy eye images to researchers.

{\bf Question 1:} Do ocular pathologies impact the enrollment process? If so, presence of which structural impairments translates into an increase in Failure To Enroll rate (FTE), \ie, the proportion of samples that could not be enrolled to the overall number of samples?

{\bf Question 2:} Does the iris recognition perform worse for non-healthy eyes with no visible impairments when photographed in near infrared (NIR) light, when compared to the healthy irides? In other words, can we assume that there are some properties of the iris image, not revealed in near infrared, that prevent iris recognition from achieving optimal performance?
  
{\bf Question 3:} What kind of visible impairments in non-healthy irides impact the iris recognition to the highest extent?

{\bf Question 4:} What are the main reasons for a bad performance when iris recognition is applied to non-healthy eyes?

To answer these questions a dataset of iris images representing more than twenty different eye diseases was built with the use of a professional iris recognition camera operating in NIR light, along with an ophthalmological commentary (Sec. \ref{sec:Database}). Most of the NIR samples are accompanied by color images to make independent ophthalmological interpretations possible. Experimental study done for three different iris recognition methods is presented in Sec. \ref{sec:Experiment}. To our knowledge, this paper offers the largest published dataset of NIR and color images for non-healthy eyes with a professional, ophthalmological commentary, and the most extensive study of how different groups of diseases may influence iris recognition.

\section{Related work}
\label{sec:RelatedWork}

Roizenblatt \emph{et al.} \cite{Roizenblatt} were probably the first to analyze how eye pathologies might influence the performance of iris recognition. They report False Non-Match Rate (FNMR) equal to 11\% when post-cataract-surgery iris images are compared to those obtained beforehand. A deterioration in FNMR in such scenario is also reported by Seyeddain \emph{et al.} \cite{Seyeddain2014}. Trokielewicz \emph{et al.} \cite{TrokielewiczWilga2014} report worse performance for three different iris recognition methods when pre-surgery, cataract-affected eyes are used in the system instead of healthy ones. Yuan \emph{et al.} \cite{Yuan} claim that laser procedures used to ablate the corneal tissue to compensate for refractive pathologies such as myopia, hypermetropia and astigmatism have little effect on iris recognition accuracy. Aslam \emph{et al.} \cite{Aslam} present the first known to us work examining biometric performance in relation to more than one eye condition, including corneal and scleral pathologies, glaucoma, scleritis and conjunctivitis. A single iris recognition methodology used by the authors turned out to be resilient for most diseases except the anterior uveitis (iritis), in which case the authors report FNMR=21\%. McConnon \emph{et al.} \cite{McConnon2012} examine the iris segmentation process (yet not the iris matching) for color images acquired by an ophthalmoscope, reporting variations by more than 2 pixels from the ''ground truth'' obtained by manual segmentation in about half of images. Borgen \emph{et al.} \cite{Borgen} build a synthetic set gathering digitally modified UBIRIS samples so that to resemble selected changes to the eye structures such as keratitis, corneal infiltrates, blurring and dulling of the cornea, corneal scarring and surgery, angiogenesis, iridectomy, iris depigmentation, tumors and melanoma. High false non-match rates (32.8\% -- 86.8\%) are reported for all modifications except for the pathological vascularization (6.6\%), changes in iris color (0.5\%) and iridectomy (0\%). It is worth noticing that most of the presented studies use a single iris recognition methodology when generating final conclusions, and all of them are based on small datasets, acquired from no more than 100 subjects.

\section{Database of iris images}
\label{sec:Database}

\subsection{Eye pathologies}
The iris, located inside the eyeball, is separated from the outside environment by the eye’s protective system -- the eyelids -- also known as palpebra, the cornea covered with tear film and the aqueous humor which fills the enclosed cavity between the cornea and the iris. In a state of normal anatomy -- open palpebral fissure, transparent cornea and clean anterior chamber of the eye -- undisturbed observation of the iridial surface is possible. However, temporary or permanent changes in the anterior part of the eye may inhibit obtaining iris images of appropriate quality. In this subsection we discuss ophthalmic pathologies represented in the database, which may prevent the use of iris biometrics technologies in practice, in respect to parts of the eye they impact most.

\textit{The cornea.} Chemical injury can deal extensive damage to the ocular surface epithelium, the cornea, and the anterior segment of the eye. It can lead to \textbf{opacification and neovascularization of the cornea}, formation of a \textbf{symblepharon and cicatricial ectropion or entropion}. If significant \textbf{corneal scarring} is present, a corneal transplant may be required to restore vision. \textbf{Pterygium} -- a benign growth of the conjunctiva, commonly forms from the nasal side of the sclera to the center of the cornea. This fibrovascular proliferation often occludes a part of the iris. \textbf{Bacterial keratitis} is an erosion or an open sore in the outer layer of the cornea with stromal infiltration, edema and hypopyon. Common pathogens that may lead to \textbf{corneal ulcers} include: Streptococcus pyogenes, Acanthamoeba, Herpes simplex, or fungal infections mainly caused by use of non-sterilized contact lenses. \textbf{Acute glaucoma} with increased pressure inside the eye occurs suddenly when the iris is pushed or pulled forward. High intraocular pressure produces symptoms such as \textbf{corneal edema, shallowness of the anterior chamber and dilatation of the pupil} which may become oval in shape. \textbf{Corneal laceration} usually requires placement of corneal sutures.

\textit{The anterior chamber.} \textbf{Hyphema} is a condition in which blood is present in the anterior chamber of the eye and may partially or completely obstruct the view of the iris. Hyphemas are frequently caused by injuries but may also occur spontaneously. A long-standing hyphema may result in hemosiderosis and heterochromia in form of partial changes in the coloration of the iris. \textbf{Hypopyon} is a leukocytic exudate present in the anterior chamber of the eye -- partially obstructing the iris, usually accompanied by redness of the conjunctiva. It is a sign of an iridial inflammation.

\textit{The iris.} \textbf{Rubeosis iridis} is a medical condition of the iris in which new, abnormal blood vessels are found on the surface of the iris. It is usually associated with disease processes in the retina. \textbf{Iris sphincter tear} is a frequent concomitant of both laceration and blunt trauma of the anterior segment. \textbf{Iridodialysis} is defined as a rupture of the iris at its thinnest area, the iris root, manifested as a separation or tearing of the iris from its attachment to the ciliary body. It is usually caused by blunt trauma to the eye, but may also be caused by penetrating eye injuries or as a complication of an intraocular surgery. Iridodialyses can be often repaired using suturing techniques. \textbf{Synechiae} are adhesions between the iris and other structures in the eye. \textbf{Iris bombe} occurs when there is a complete adhesion (posterior synechiae) between the iris and the anterior capsule of the lens creating a 360-degree area of the adhesion.

\textit{The lens.} With \textbf{anterior lens luxation}, the lens enters the anterior chamber of the eye. This can cause damage to the cornea, swelling and progression of lens opacity, so the iris image can become blurry. \textbf{Phacolytic glaucoma} is an inflammatory glaucoma caused by the leakage of lens protein through the capsule of a hyper-mature cataract. Escalating \textbf{corneal edema and milky aqueous humor} in the anterior chamber also blur the iris image.

\emph{Pars plana \textbf{vitrectomy}} is a general term used to describe a group of surgical procedures performed in the deeper part of the eye, behind the lens. Silicone oil is used as an intraocular tamponade in the repair of {\bf retinal detachment} or {\bf diabetic retinopathy}. Sometimes it may relocate itself to the front of the iris, causing a blurry iridial image.

\subsection{Data collection protocol}
For this study a new database was designed and collected specifically for the assessment of how the iris recognition is immune, or prone, to ocular pathologies. The dataset comprises images collected from patients during routine ophthalmology examinations. All patients participating in the study have been provided with detailed information on the research and a written consent has been obtained from each volunteer.

Data collection lasted approximately 16 months. During each visit, both NIR-illuminated images (compliant with the ISO/IEC 19794-6:2011 recommendations) and standard color photographs (for selected cases) were acquired to perform visual inspection of the illness in samples showing significant alterations to the eye. The data have been acquired with three commercial cameras: 1) the IrisGuard AD100 for NIR images, 2) Canon EOS1000D with EF-S 18-55 mm f/3.5-5.6 lens equipped with a Raynox DCR-250 macro converter and a ring flashlight suited for macrophotography, and 3) an ophthalmology slit-lamp camera Topcon DC3, Tab. \ref{table:database_summary}.

\begin{figure*}[!t]
\centering
\includegraphics[width=0.78\textwidth]{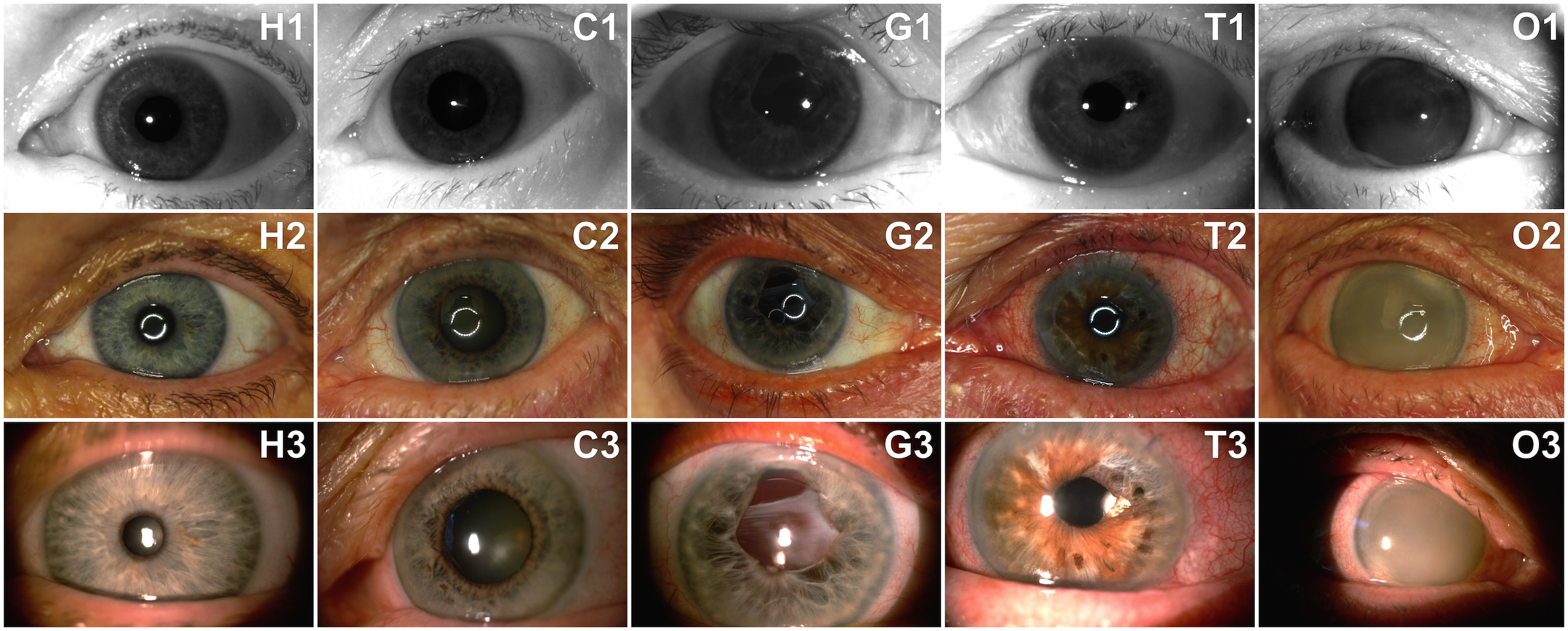}
\caption{Samples of 5 different eyes acquired using three different imaging systems: IrisGuard AD-100 (top row), Canon EOS 1000D (middle row), and Topcon DC3 slit-lamp camera (bottom row). Each column includes samples corresponding to a different group used further in our experimental study, namely: healthy eye (H1-H3), non-healthy eye but with a clear pattern (C1-C3), an eye with geometrical deviations (G1-G3), an eye with iris tissue impairments (T1-T3), and an eye with obstructions in front of the iris (O1-O3).}
\label{fig:samples}
\end{figure*}

\subsection{Database statistics}

The entire dataset comprises 2996 images of 230 distinct irides, Tab. \ref{table:database_summary}. Every class contains NIR-illuminated images, while for some of them visible spectrum photographs have also been taken (in cases when visual inspection revealed significant changes to the structures of the eye). Fig. \ref{fig:samples} shows samples of five different eyes obtained using all three devices. 

For 184 images there is one sample collection session, for 38 classes there are two sessions, for 6 classes there are three sessions, and finally for 2 classes there are four different acquisition sessions. Usually, the second and subsequent sessions contain images obtained after some kind of medical procedure, \eg, a cataract surgery. Detailed information, including precise description of medical conditions and procedures performed in each case, is disclosed in the metadata that accompany the dataset. No data censoring was performed when issuing the dataset, except for removing images that did not show an eye at all.

\begin{table}[!ht]
\renewcommand{\arraystretch}{1.1}

\caption{Formats and numbers of images collected by each device.}
\label{table:database_summary}
\centering\footnotesize
\begin{tabular}[t]{|c|c|c|}
\hline
\textbf{Device} & \textbf{Image format} & \textbf{Number of images}\\
\hline
\hline
IrisGuard AD100 & grayscale, 640x480 BMP & 1793\\
\hline
Canon EOS1000D & color, 10 Mpixel JPEG & 868\\
\hline
Topcon DC3 & color, 8 Mpixel
 JPEG & 335\\
\hline
\end{tabular}
\end{table}

\section{Experimental study}
\label{sec:Experiment}

\subsection{Dividing the data}

Ophthalmological practice shows that most non-healthy eyes suffer from more than one condition, which are often unrelated and impacting the eye in different ways. While some illnesses cause the pupil to distort and deviate from its usual circular shape, other pathologies impact the iris directly or cause changes to other parts of the eyeball, such as the uvea, the cornea, the anterior chamber, or even to the retina. Hence, conducting an insightful analysis can be challenging. The data were therefore partitioned respectfully to the type of influence that given ocular pathology inflicts on the eye. This allows us to devise five different partitions: \emph{Healthy} partition, comprising healthy eyes only, \emph{Clear} partition made up of eyes with a disease present, but not affecting the eye structures perceivably, \emph{Geometry} partition (eyes with pupil geometry distorted by the pathology), \emph{Tissue} partition (eyes with damage inflicted on the iris tissue) and \emph{Obstructions} partition, encompassing the eyes with obstructions present in front of the iris. Table \ref{table:FTE} presents numbers of classes (\ie, different eyes) and images in each category. Figure \ref{fig:samples} shows sample images belonging to each partition.

\subsection{Methodology}

{\bf To answer the first question} formulated in the introductory part, the failure-to-enroll error rates are calculated for each database partition and three different iris recognition methods. {\bf To answer questions two and three}, all possible genuine comparisons were generated and full cross comparisons were made to get all the possible impostor comparison scores for each dataset partition. One-tailed two-sample t-test at the significance level $\alpha = 0.05$ is used to compare averages of the comparison scores across partitions. That is, in all t-tests the null hypothesis states that the comparison scores calculated for samples belonging to two partitions being compared (for instance, \emph{Healthy} and \emph{Clear} samples) come from independent random samples with equal means (equality of variances is not assumed). Additionally to testing the equality of means, we run two-sample Kolmogorov-Smirnov test at the significance level $\alpha = 0.05$ to check whether samples belonging to two partitions come from the same distribution. These analyses were done independently for three iris recognition methods. Finally, the segmentation errors are analyzed and visual inspection of the eye samples resulting in the worst comparison scores are performed {\bf to answer the last, fourth question}.

\begin{table*}[!ht]
\renewcommand{\arraystretch}{1.1}
\caption{Columns 2 and 3 show the number of samples and unique irides in each dataset partition. Columns 4 through 6 present FTE rates obtained in each partition for three iris recognition methods used in this work. The worst results for each method are \textbf{bolded}.}
\label{table:FTE}
\centering\footnotesize
\begin{tabular}[t]{|c|c|c|c|c|c|}
\hline
\textbf{Data partition} & \textbf{Number of irides} & \textbf{Number of images} & {FTE for MIRLIN} & {FTE for VeriEye} & {FTE for OSIRIS} \\
\hline
\hline
\textbf{Healthy} & 35 & 216 & 1.85\% & 0\% & 0\% \\
\hline
\textbf{Clear} & 87 & 568 & 4.40\%  & 0	\% & 1.23\% \\
\hline
\textbf{Geometry} & 53 & 312 & 16.03\% & \textbf{5.13\%} & 5.45\% \\
\hline
\textbf{Tissue} & 8 & 50 & 2\% & 0\% & 0\% \\
\hline
\textbf{Obstructions} & 36 & 207 & \textbf{18.36\%}  &  3.86\% & \textbf{8.21\%} \\
\hline
\end{tabular}
\end{table*}

\subsection{Iris recognition methods used in this research}

Three independent iris matching methods are used in this work and briefly characterized in this subsection. {\bf MIRLIN} ({\it Monro Iris Recognition Library and INterface}) is a commercial implementation of the concept proposed by Monro \etal \cite{Monro2007}. In this approach the iris features are derived from zero-crossings of the differences between Discrete Cosine Transform (DCT) calculated in rectangular iris image patches. The resulting iris binary feature sets are compared by calculating fractional Hamming distance. Two matching irides should have this distance close to zero, while 0.5 is expected when two different eyes are being matched. {\bf OSIRIS} ({\it Open Source for IRIS}) follows a well-known iris recognition concept proposed by Daugman and includes an iris image normalization into a dimensionless polar representation, Gabor-based filtering, quantization of the filtering results into a binary iris code, and code matching based on fractional Hamming distance \cite{OSIRIS}. As in MIRLIN method, close-to-zero comparison scores are expected when same-eye samples are compared, and 0.5 should be obtained for non-same-eye images. Details of the third method, {\bf VeriEye}, are not revealed by the vendor, apart from the claim of using non-circular approximations of the iris and pupil boundaries, and coding which does not follow Gabor filtering \cite{VeriEye}. The comparison score starts from zero (when totally different eyes are compared) to some, yet unknown high values when same-eye samples are compared.

\section{Results}
\label{sec:results}

\subsection{Enrollment performance (re: Question 1)}


FTE rates  obtained in each partition (Tab. \ref{table:FTE}) suggest that iris recognition performs particularly bad for samples included in \emph{Geometry} and \emph{Obstructions} partitions. Those partitions comprise images with pupil either distorted or not visible at all due to various types of occlusions. Hence, {\bf the answer to the question 1 is that the enrollment stage is highly sensitive to those conditions that make the pupil geometry distorted, or the iris pattern partially obstructed}.

\subsection{Matching performance (re: Questions 2 and 3)}

\begin{figure*}[!htb]
\centering
\includegraphics[width=0.33\textwidth]{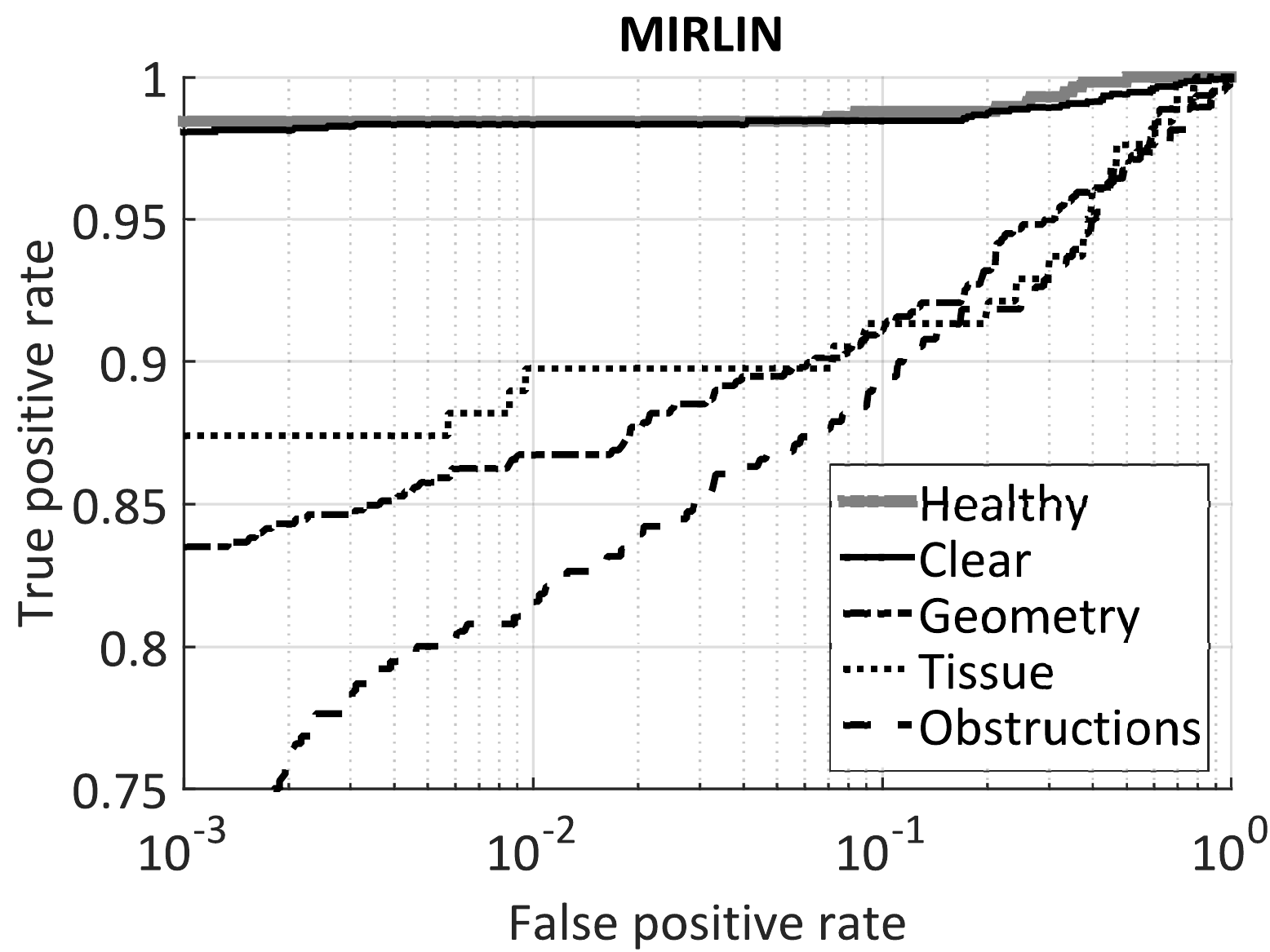}\hfill
\includegraphics[width=0.33\textwidth]{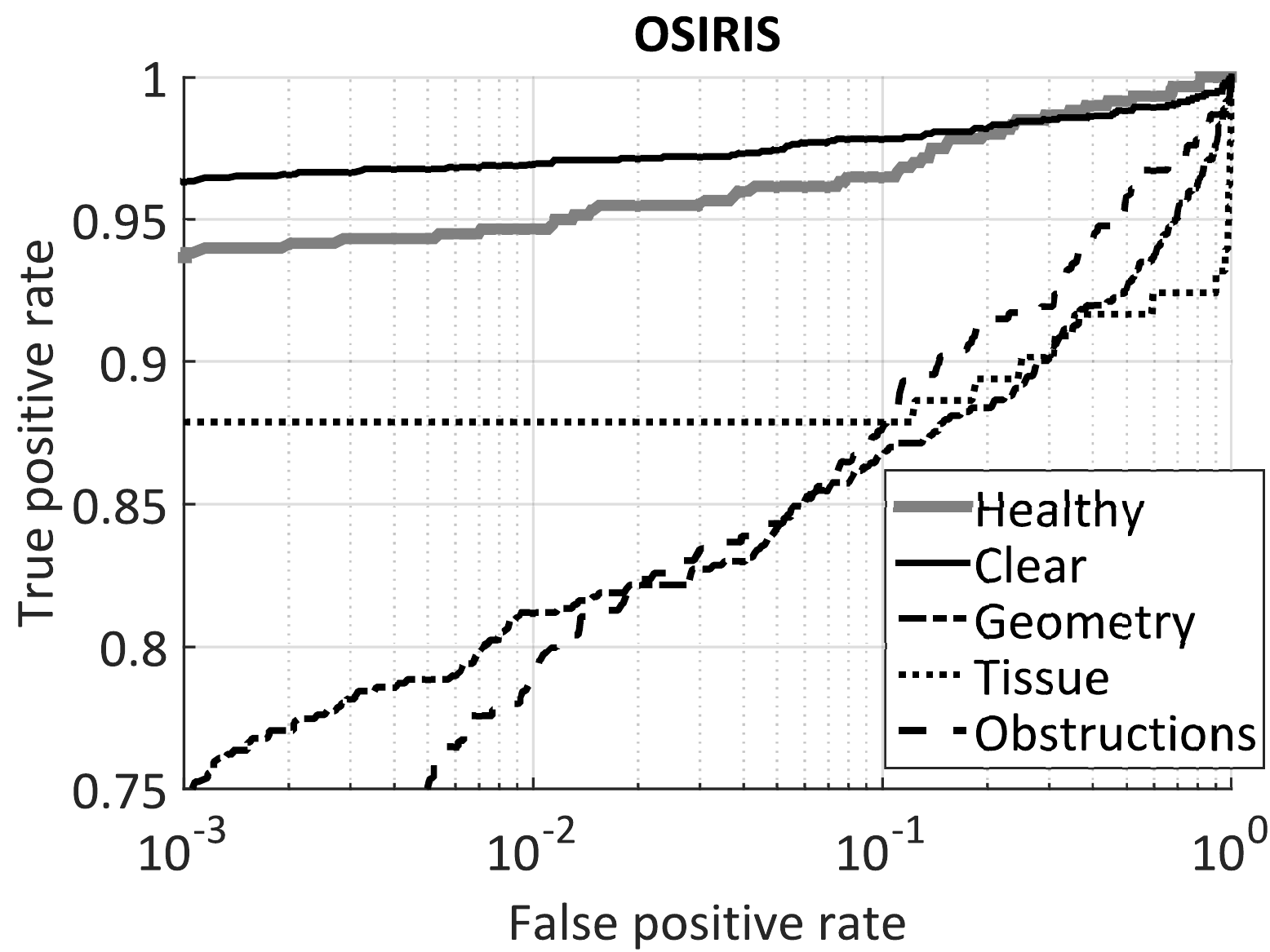}\hfill
\includegraphics[width=0.33\textwidth]{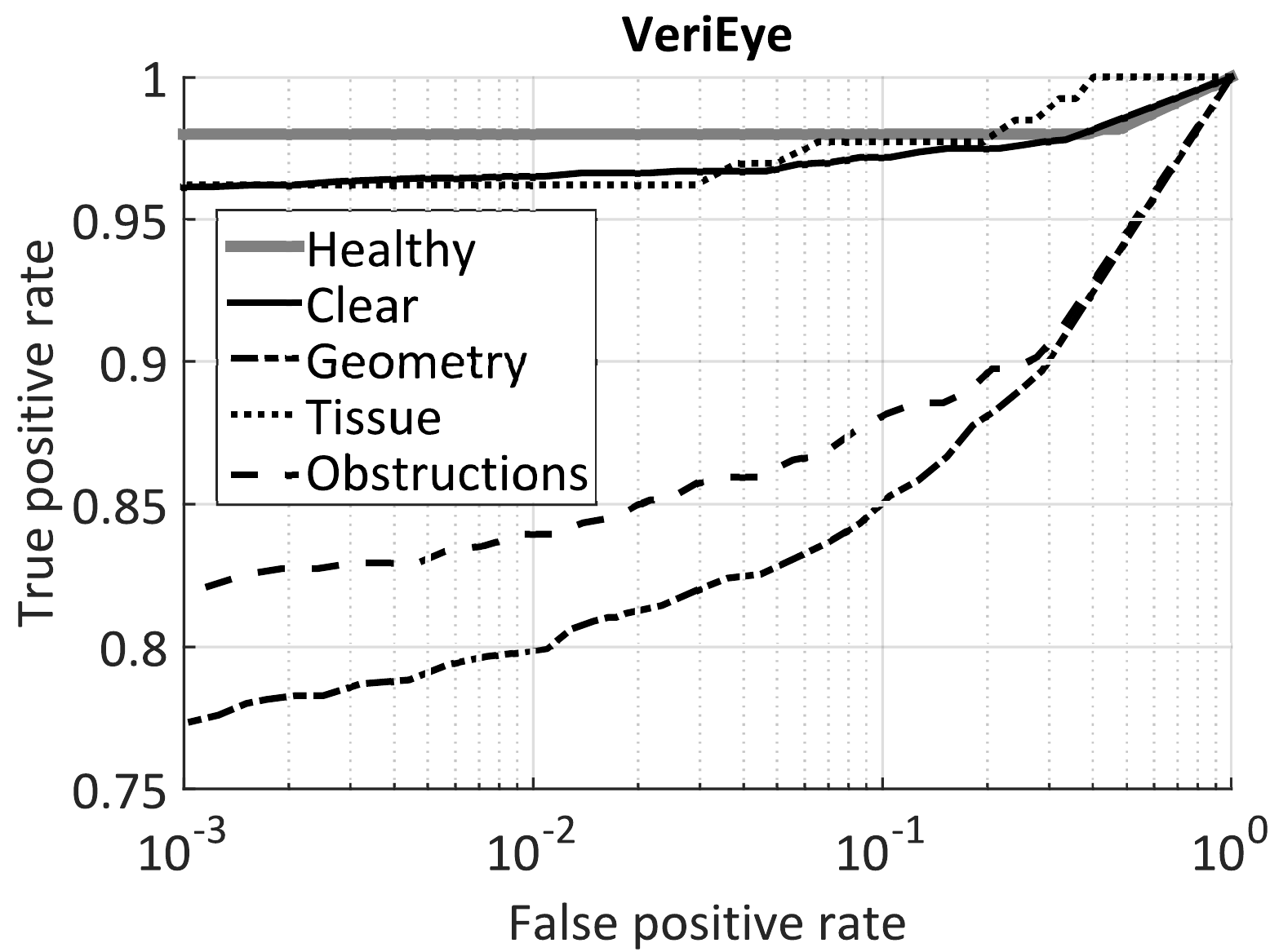}
\caption{Receiver operating characteristics (ROC) for MIRLIN (left), OSIRIS (middle) and VeriEye (right) iris recognition methods.}
\label{fig:CDF}
\end{figure*}

To check whether the iris recognition performs worse for non-healthy eyes (although not revealing visible changes) than healthy eyes (question 2), the genuine and impostor scores for \emph{Healthy} and \emph{Clear} eyes are compared (Fig. \ref{fig:CDF}, solid lines) for those irides that were correctly enrolled. 

For all iris coding methodologies we observe slightly worse mean genuine comparison scores in \emph{Clear} partition when compared to the \emph{Healthy} eyes. Mean values of the impostor scores differ not too much. Although t-test suggests that those differences in average of genuine scores are statistically insignificant for MIRLIN and OSIRIS methods ($p$-value $ > \alpha = 0.05$), the Kolmogorov-Smirnov test points at statistically significant differences in distributions for all tested methods and both for the genuine and impostor comparison scores, Tab. \ref{table:statTests}. Hence, {\bf the answer to the question 2 is affirmative, but the observed differences are uneven across the methods and small for impostor comparison scores}. The iris recognition methods based on image filtering (MIRLIN and OSIRIS) seem to be more robust to those eye conditions that make no visible changes in the iris structure.

To answer the third question, related to how diseases introducing visible structural changes to the iris pattern impact the performance, the genuine and impostor comparison scores were calculated in the following three partitions of non-healthy eyes: \emph{Geometry}, \emph{Tissue} and \emph{Obstructions}, Fig. \ref{fig:CDF}, dashed and dotted lines. In this experiment, we see serious deterioration in within-class matching performance. Both t-test and Kolmogorov-Smirnov test give statistically significant differences for the genuine and impostor scores across the partitions, but for the MIRLIN method (cf. Table \ref{table:statTests}). The most influential to all iris recognition methods are conditions generating obstructions of the iris structure. Observing the mean values of the genuine comparison scores for all the applied iris recognition methods, it is evident that such a large shift in genuine comparison distributions would have a significant influence on a false-non match rate, difficult to compensate by modification of the acceptance threshold. However, observed shifts of the impostor score distributions are small and have rather marginal influence on the overall system performance. Hence, answering the question 3, we state that {\bf all eye conditions resulting in visible iris structure impairments have a substantial influence on within-class variability. The largest increase in the probability of a false non-match is expected for those conditions that make geometrical deformations of the pupil and introduce obstructions of the iris.}

\newlength{\colw}
\setlength{\colw}{0.965cm}

\begin{table*}[!htb]
\renewcommand{\arraystretch}{1.1}

\caption{Summary of statistical testing of differences in average comparison scores obtained for all iris recognition methods in five partitions of the dataset. In {\bf t-test} the null hypotheses state that the comparison scores calculated for samples belonging to two partitions being compared come from distributions of the same mean. Alternative hypotheses are defined separately for each case and they are shown in rows labeled `H1 (t-test)'.  In {\bf Kolmogorov-Smirnov test} the null hypotheses state that the data of two compared partitions are from the same distribution. Alternative hypotheses state the opposite and are detailed in rows labeled `H1 (KS-test)'. In both tests, $p$-values not exceeding $\alpha=0.05$ denote that the null hypothesis was rejected and the corresponding alternative hypothesis was selected. $g$ and $i$ denote the genuine and impostor scores, correspondingly, while $\bar{g}$ and $\bar{i}$ denote their means. The sign $\nsim$ denotes inequality of distributions.}

\label{table:statTests}
\centering\scriptsize
\begin{tabular}[t]{cc|>{\centering}m{\colw}|>{\centering}m{\colw}|>{\centering}m{\colw}|>{\centering}m{\colw}|c||>{\centering}m{\colw}|>{\centering}m{\colw}|>{\centering}m{\colw}|>{\centering}m{\colw}|c|}
\cline{3-12}
& & \multicolumn{5}{c||}{\bf Genuine comparison scores} & \multicolumn{5}{c|}{\bf Impostor comparison scores} \\
\cline{3-12}
& & \emph{Healthy} & \emph{Clear}  & \emph{Geometry} & \emph{Tissue} & \emph{Obstr.}  & \emph{Healthy} & \emph{Clear}  & \emph{Geometry} & \emph{Tissue} & \emph{Obstr.} \\
& & ($g_h$) & ($g_c$)  & ($g_g$) & ($g_t$) & ($g_o$) & ($i_h$) & ($i_c$)  & ($i_g$) & ($i_t$) & ($i_o$) \\\hline\hline
\multicolumn{1}{|c}{\multirow{3}{*}{MIRLIN}} & \multicolumn{1}{|c|}{\bf mean} & 0.0198 & 0.0236 & 0.0897 & 0.0773 & 0.1080 & 0.4053 & 0.4027 & 0.4050 & 0.4158 & 0.4120\\\cline{2-12}
\multicolumn{1}{|c}{} & \multicolumn{1}{|c|}{\bf H1 (t-test)} & & $\bar{g}_c > \bar{g}_h$ & $\bar{g}_g > \bar{g}_h$ & $\bar{g}_t > \bar{g}_h$ & $\bar{g}_o > \bar{g}_h$ & & $\bar{i}_c < \bar{i}_h$ & $\bar{i}_g < \bar{i}_h$ & $\bar{i}_t > \bar{i}_h$ & $\bar{i}_o > \bar{i}_h$ \\\cline{2-12}
\multicolumn{1}{|c}{} & \multicolumn{1}{|c|}{\bf $p$-value (t-test)} & & 0.0666 & \texttildelow 0 &  \texttildelow 0 &  \texttildelow 0 & & \texttildelow 0 & 0.2452 & \texttildelow 0 & \texttildelow 0 \\\cline{2-12}
\multicolumn{1}{|c}{} & \multicolumn{1}{|c|}{\bf H1 (KS-test)} & & $g_c\nsim g_h$ & $g_g\nsim g_h$ & $g_t\nsim g_h$ & $g_o\nsim g_h$ & & $i_c\nsim i_h$ & $i_g\nsim i_h$ & $i_t\nsim i_h$ & $i_o\nsim i_h$ \\\cline{2-12}
\multicolumn{1}{|c}{} & \multicolumn{1}{|c|}{\bf $p$-value (KS-test)} & & 0.0178 & \texttildelow 0 & \texttildelow 0 & \texttildelow 0 & & \texttildelow 0 & \texttildelow 0 & 0.0002 & \texttildelow 0 \\\hline\hline
\multicolumn{1}{|c}{\multirow{3}{*}{OSIRIS}} & \multicolumn{1}{|c|}{\bf mean} & 0.2424 & 0.2483 & 0.3084 & 0.2701 & 0.3100 & 0.4688 & 0.4656 & 0.4675 & 0.4713 & 0.4700\\\cline{2-12}
\multicolumn{1}{|c}{} & \multicolumn{1}{|c|}{\bf H1 (t-test)} & & $\bar{g}_c > \bar{g}_h$ & $\bar{g}_g > \bar{g}_h$ & $\bar{g}_t > \bar{g}_h$ & $\bar{g}_o > \bar{g}_h$ & & $\bar{i}_c < \bar{i}_h$ & $\bar{i}_g < \bar{i}_h$ & $\bar{i}_t > \bar{i}_h$ & $\bar{i}_o > \bar{i}_h$ \\\cline{2-12}
\multicolumn{1}{|c}{} & \multicolumn{1}{|c|}{\bf $p$-value (t-test)} & & 0.0556 & \texttildelow 0  & 0.0012  & \texttildelow 0  & & \texttildelow 0 & \texttildelow 0 & \texttildelow 0 & \texttildelow 0 \\\cline{2-12}
\multicolumn{1}{|c}{} & \multicolumn{1}{|c|}{\bf H1 (KS-test)} & & $g_c\nsim g_h$ & $g_g\nsim g_h$ & $g_t\nsim g_h$ & $g_o\nsim g_h$ & & $i_c\nsim i_h$ & $i_g\nsim i_h$ & $i_t\nsim i_h$ & $i_o\nsim i_h$ \\\cline{2-12}
\multicolumn{1}{|c}{} & \multicolumn{1}{|c|}{\bf $p$-value (KS-test)} & & \texttildelow 0 & \texttildelow 0 & 0.0024 & \texttildelow 0 & & \texttildelow 0 & \texttildelow 0 & 0.0009 & \texttildelow 0 \\\hline\hline
\multicolumn{1}{|c}{\multirow{3}{*}{VeriEye}} & \multicolumn{1}{|c|}{\bf mean} & 500.18 & 458.50 & 282.28 & 447.92 & 265.62 & 3.479 & 2.998 & 1.799 & 2.598 & 2.088 \\\cline{2-12}
\multicolumn{1}{|c}{} & \multicolumn{1}{|c|}{\bf H1 (t-test)} & & $\bar{g}_c < \bar{g}_h$ & $\bar{g}_g < \bar{g}_h$ & $\bar{g}_t < \bar{g}_h$ & $\bar{g}_o < \bar{g}_h$ & & $\bar{i}_c < \bar{i}_h$ & $\bar{i}_g < \bar{i}_h$ & $\bar{i}_t < \bar{i}_h$ & $\bar{i}_o < \bar{i}_h$ \\\cline{2-12}
\multicolumn{1}{|c}{} & \multicolumn{1}{|c|}{\bf $p$-value (t-test)} & & \texttildelow 0 & \texttildelow 0 & 0.001 & \texttildelow 0 & & \texttildelow 0 & \texttildelow 0 & \texttildelow 0 & \texttildelow 0 \\\cline{2-12}
\multicolumn{1}{|c}{} & \multicolumn{1}{|c|}{\bf H1 (KS-test)} & & $g_c\nsim g_h$ & $g_g\nsim g_h$ & $g_t\nsim g_h$ & $g_o\nsim g_h$ & & $i_c\nsim i_h$ & $i_g\nsim i_h$ & $i_t\nsim i_h$ & $i_o\nsim i_h$ \\\cline{2-12}
\multicolumn{1}{|c}{} & \multicolumn{1}{|c|}{\bf $p$-value (KS-test)} & & \texttildelow 0 & \texttildelow 0 & 0.0173 & \texttildelow 0 & & \texttildelow 0 & \texttildelow 0 & 0.0002 & \texttildelow 0 \\\hline
\end{tabular}
\end{table*}

\subsection{Sources of errors (re: Question 4)}

To seek for probable sources of erratic performance, sample pairs yielding genuine match scores below typical acceptance thresholds (\ie, $0.32$ for MIRLIN and OSIRIS, and $30$ for VeriEye) were visually inspected. The most likely cause or errors is a failed segmentation that causes the algorithm to encode non-iris image portions. These segmentation errors were caused mostly by either irregular pupil boundary, corneal occlusions that obstruct the pupil and the iris, or by damage to the iris tissue being misinterpreted by the pupil segmentation algorithms as the pupil itself.  In particular, all of the sample pairs generating the worst OSIRIS scores (\emph{Geometry} and \emph{Obstructions}) devise their poor performance from the segmentation errors. This is similar when the MIRLIN matcher is involved, except for two images that produce poor scores in the \emph{Clear} subset: one of which is blurred, and the other showing no particular flaw. For the VeriEye matcher there is no way of assessing if segmentation is done correctly, however, when looking at those samples that perform the worst, one may identify conditions responsible for errors, namely: significant geometrical distortions, severe corneal hazes, blurred boundary between the iris and the pupil. Thus, segmentation failures may be the case here as well.

\section{Conclusions}

This paper provides the most thorough analysis of iris recognition performance in the presence of various ocular pathologies to date. The fact that usually a few different and unrelated diseases might appear in a single eye, requires a novel approach based not on a disease taxonomy, but on the type of damage inflicted on the eye. Decrease in the performance begins manifesting itself as early as in the enrollment phase, where higher FTE rates are observed for eyes with geometrical distortions in the pupillary area and for those with corneal occlusions. Following that, eyes with those pathologies perform far worse (mostly in terms of the genuine comparison scores) than their healthy counterparts. For all iris coding methods there are also statistically significant (according to the Kolmogorov-Smirnov test) small differences in comparison scores calculated for healthy eyes and for non-healthy ones, but not affected perceivably in a visual inspection. In most cases the erroneous segmentation is the main reason of a decreased performance. The fact that the worse degradation in recognition accuracy involves eyes with severely changed internal structures lets us hope that a simple visual inspection of the eye (performed \eg, by the person supervising the enrollment) could prevent most errors. In such cases, persons with diseased eyes may be encouraged to enroll using the second eye (if healthy), or a different biometric characteristic (if applicable). It is worth noting that the database collected for this research is available to the biometric community for non-commercial research purposes at no cost\footnote{Further information on how to get the data can be found at: \\\url{http://zbum.ia.pw.edu.pl/EN}$\rightarrow$Research$\rightarrow$Databases}. 

{\small
\bibliographystyle{ieee}
\bibliography{refs}
}

\end{document}